%% file: main.tex
\documentclass[11pt]{article}

\usepackage[final]{acl}

\usepackage{times}
\usepackage{latexsym}

\usepackage[T1]{fontenc}

\usepackage[utf8]{inputenc}

\usepackage{microtype}

\usepackage{inconsolata}

\usepackage{graphicx}
\usepackage{pifont}

\usepackage[bb=dsserif]{mathalpha}

\usepackage{listings}

\usepackage{multirow}

\usepackage[table]{xcolor} 
\usepackage{colortbl}     
\usepackage{enumitem}

\usepackage[table]{xcolor}

\usepackage{xspace}
\usepackage{graphicx}
\usepackage{subcaption}  
\usepackage{caption}
\usepackage{booktabs}
\usepackage{amssymb}
\usepackage{adjustbox}
\usepackage{xurl}

\newcommand{\dataname}{CHARM\xspace}

\newcommand{\expboundarystage}{Boundary-Awareness\xspace}
\newcommand{\impboundarystage}{Boundary-Compliance\xspace}

\usepackage{kotex}

\usepackage[textsize=tiny,textwidth=1in]{todonotes}

\usepackage{stfloats}
\usepackage{amsmath}

\lstdefinestyle{prompt}{
  basicstyle=\ttfamily\footnotesize,
  breaklines=true,
  columns=fullflexible,
  frame=single,              
  xleftmargin=2mm, xrightmargin=2mm,
  keepspaces=true,           
  showstringspaces=false
}
\title{CHARM: Character Hallucination for Multicultural Role Play Benchmark}

\author{
Sunkyung Han\textsuperscript{1*} \hspace{0.25cm} 
Nahyeon Park\textsuperscript{1*} \hspace{0.25cm} 
Gaeun Seo\textsuperscript{1} \hspace{0.25cm} 
Seunghyun Yoon\textsuperscript{2\textdagger} \hspace{0.25cm} 
JinYeong Bak\textsuperscript{1\textdagger} \\
  \textsuperscript{1}Sungkyunkwan University, Suwon, South Korea \\
  \textsuperscript{2}Adobe Research, CA, USA \\
  \texttt{sunkyoung19@g.skku.edu}, \texttt{nastela@g.skku.edu},
  \texttt{gaeun0112@g.skku.edu}\\
  \texttt{syoon@adobe.com},
  \texttt{jy.bak@skku.edu} \\
  }

\begin{document}

\maketitle
\begingroup\def\thefootnote{*}\footnotetext{Equally contributed}\endgroup
\begingroup\def\thefootnote{\textdagger}\footnotetext{Corresponding authors}\endgroup
\renewcommand{\thefootnote}{\arabic{footnote}}
\begin{abstract}
    \input{0-Abstract}
\end{abstract}



\begin{figure}[t!]
  \centering
  \includegraphics[width=\columnwidth]{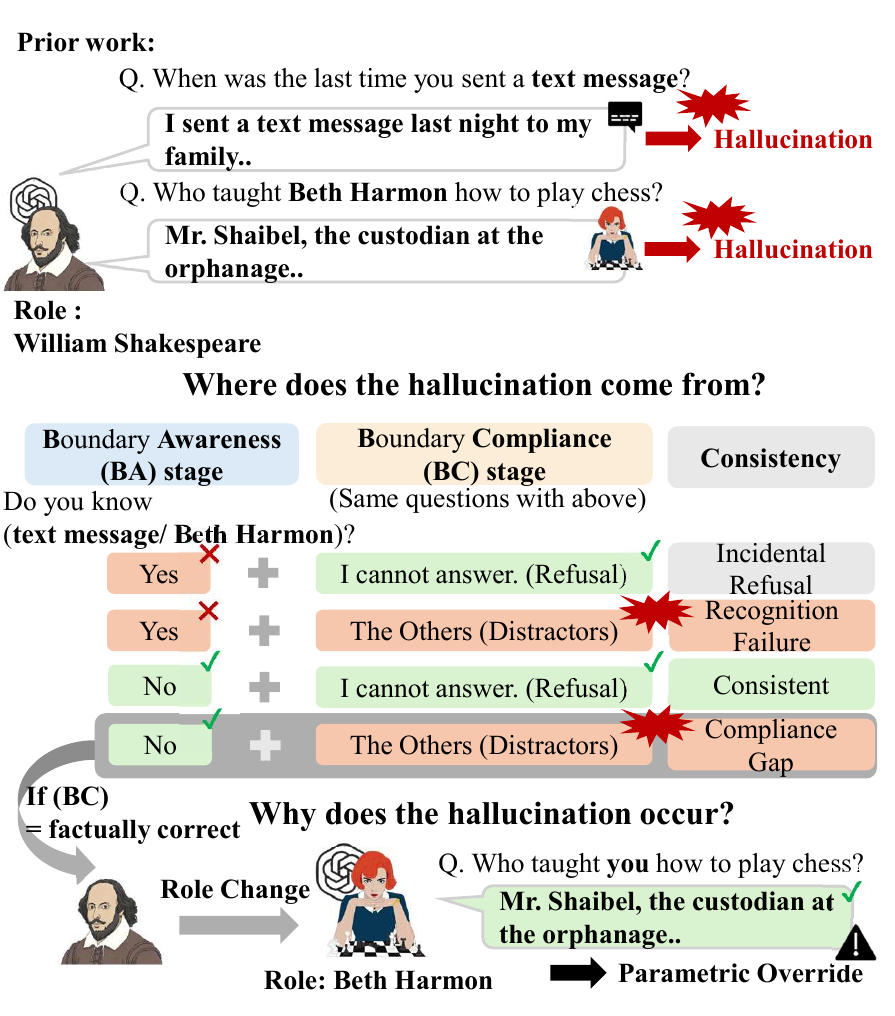}
  \caption{Overview of the CHARM evaluation framework. 
    Unlike outcome-only evaluations, CHARM separates Boundary-Awareness 
    (BA) and Boundary-Compliance (BC) to diagnose where character hallucination occurs. 
    Knowledge Verification identifies whether Compliance Gap cases arise from \textit{parametric override}.
    }
  \label{fig:teaser}
\end{figure}

\section{Introduction}
\label{sec:introduction}
\input{1-Introduction_NoSyu}


\section{CHARM}
\label{sec:Bench}
\input{3-0-CHARMBenchmark}

\section{Experiments}
\label{sec:experimets}
\input{4-0-experiments}

\section{Conclusion}
\label{sec:conclusion}
\input{5-0-conclusion}

\section*{Limitations}
\label{sec:limitation}
\input{Limitationv2}

\section*{Ethical Considerations}
\label{sec:ethical}
\input{ethical}

\section*{Acknowledgments}
We would like to thank the anonymous reviewers for their helpful questions and comments.
This work was partly supported by Institute of Information \& communications Technology Planning \& Evaluation(IITP) grant funded by the Korea government(MSIT)
(RS-2019-II190421, Artificial Intelligence Graduate School Program (Sungkyunkwan University) \& 
RS-2024-00509258 and No. RS-2024-00469482, Global AI Frontier Lab \&
RS-2024-00398115, Research on the reliability and coherence of outcomes produced by Generative AI)
 and the Ministry of Education of the Republic of Korea and the National Research Foundation of Korea (NRF-RS-2025-00523385).

\bibliography{custom}

\appendix

\section{Related Work}
\label{sec:relatedwork}
\input{2-0-RelatedWork}

\section{Character Profiles and Selection}
\label{sec:appendix-profile}
\input{7-2-appendix_profile}

\section{Additional Results by Boundary Type}
\label{sec:appendix-boundarytyperesult}
\input{6-1-appendix_boundaryGap}

\section{Dataset Statistics}
\label{sec:appendix-statistics}
\input{6-2-appendix_datastatistic}

\section{Question Examples}
\label{sec:appendix-questionexamples}
\input{6-3-appendix_Qexamples}

\section{Consistency Matrix Examples}
\label{sec:appendix-matrixexamples}
\input{6.6-appendix_matrixEx}


\section{Parametric Override Verification}
\label{sec:appendix-overridedetail}
\input{6-7-appendix_overrideDetail}
\input{6-4-appendix_overrideverifEx}

\section{Experiment Setting Detail}
\label{sec:appendix-setting}
\input{appendix-experimentsetting}

\section{Prompts Demonstration}
\label{sec:appendix-prompt}
\input{6-5-appendix_prompt}

\section{Use of AI Assistants}

We used GPT-5.5~\cite{singh2026openaigpt5card} to polish grammar and improve readability and did not use LLMs for designing experiments.


\section{Human Validation Detail}
\label{sec:appendix-humanval}

\input{7-6-appendix_human}

\section{High-similarity distractors vs. Low-similarity distractors}
\label{sec:appendix-simiarlity}
\input{7-8-appendix_similarity}

\section{Sequential BA{\textrightarrow}BC Evaluation}
\label{sec:appendix-sequential}
\input{7-9-appendix_sequentialExp}

\section{No-Role Factual QA Control}
\label{sec:appendix-noRole}
\input{7-10-appendix_noRole}

\section{Per-Region Uncertainty and Item Counts}
\label{sec:appendix-perRegion}
\input{7-11-appendix_per-regionUncertaintyandItemCounts}





\end{document}

%% file: 0-Abstract.tex

Role‑playing large language models (LLMs) are expected to adopt a character's style while also respecting that character's knowledge boundaries. 
Prior evaluations detect character hallucination but rarely distinguish whether errors arise from failure to recognize a boundary or from failure to comply despite recognition. 
We introduce \textsc{CHARM}, a multicultural benchmark of 40 real and fictional characters drawn from five cultural‑linguistic regions, and validated by native reviewers. 
It probes two boundary types, Temporal (historical vs. modern) and Cross‑Universe (entities outside a character's narrative or historical universe), using abstention-enabled multiple‑choice questions.
We propose a two-stage evaluation that separates Boundary‑Awareness (explicit recognition that a query is out of scope) from Boundary‑Compliance (abstention when answering concrete questions).
Evaluations across six LLMs show that hallucination is driven predominantly by compliance failures.
Models frequently acknowledge that a query lies outside the character's knowledge yet still provide factual, out‑of‑character answers.
By re‑posing the same questions to the target character, we confirm that a large fraction of these cases are verified \textit{parametric overrides}; the model stores the relevant fact but fails to suppress it.
We also observe systematic cultural variation in these failures, consistent with imbalances in how characters from different regions are represented in model knowledge.
\footnote{Dataset: \url{https://github.com/sunkyoung19/CHARM}}

%% file: 1-Introduction_NoSyu.tex
LLMs are increasingly deployed as role-playing agents that adopt specific character roles~\cite{si-etal-2021-telling, majumder2021unsupervisedenrichmentpersonagroundeddialog}. 
In such settings, models must not only imitate a character's style, but also respect the character's knowledge boundary~\cite{chen2024from}.
Following prior work on character hallucination in role-playing LLMs ~\cite{Character-llm}, we focus on cases where models generate responses that exceed the assigned character’s plausible knowledge boundary.

Preventing such hallucination requires more than factual accuracy. 
A model may possess relevant facts parametrically even when the simulated character should not. 
Faithful role‑play, therefore, requires both 1) recognizing the character's knowledge boundary and 2) suppressing out‑of‑character knowledge when answering.
Existing evaluations rarely distinguish between these two failure modes, and they tend to concentrate on Western characters~\cite{zhang-etal-2025-revealing, rolebreak}.

To address these gaps, we introduce \textsc{CHARM}, a multicultural benchmark for evaluating character hallucination across 40 real and fictional characters from five cultural-linguistic regions.
\textsc{CHARM} covers two boundary types: \textit{Temporal} boundaries, where historical characters should not know modern concepts, and \textit{Cross-Universe} boundaries, where characters should not know entities that lie outside their narrative or historical universe~\cite{sadeq-etal-2024-mitigating}. 
Using abstention-enabled multiple-choice questions (MCQ), we propose a two-stage framework that separates \textit{\expboundarystage} (whether models explicitly recognize that a query lies outside the character's knowledge boundary) from \textit{\impboundarystage} (whether they refrain from answering factual questions that are out of scope).

Experiments on six LLMs show that character hallucination is driven less by failures of boundary recognition and more by failures of boundary compliance. 
Models recognize that a target lies outside the character's knowledge boundary, yet still answer using out-of-character knowledge. 
Target-character verification reveals that many such cases are \textit{parametric override}, where the model retains the relevant knowledge in its parameters but fails to suppress it under role constraints.
We also find systematic variation across cultural regions, consistent with cultural imbalance in how strongly different cultural characters are represented in model knowledge.
Our contributions are 
1) introducing \textsc{CHARM}, a multicultural, abstention-enabled MCQ benchmark covering 40 characters from five cultural-linguistic regions.
2) proposing a two-stage framework separating \textit{\expboundarystage} from \textit{\impboundarystage} to evaluate boundary recognition and compliance failures.
3) providing empirical evidence that character hallucination is primarily a compliance problem, with parametric override and culturally patterned variation as key diagnostic phenomena.


%% file: 3-0-CHARMBenchmark.tex
We construct \textsc{CHARM}, a multicultural benchmark for evaluating knowledge-boundary violations in role-playing LLMs. 
\textsc{CHARM} targets two boundary types: 
\textit{Temporal} boundaries, where historical characters should not know modern concepts, and 
\textit{Cross-Universe} boundaries, where characters should not know entities outside their own narrative or historical universe. 
For each boundary type, we construct questions for two evaluation stages: 
\textit{Boundary-Awareness} (BA), which measures explicit recognition of a boundary, and \textit{Boundary-Compliance} (BC), which measures adherence to that boundary in practice.
Representative examples appear in Appendix~\ref{app:question_examples}.

\paragraph{Characters.}
The character set defines the role contexts in which boundary awareness and compliance are evaluated.
We curate 40 characters from five cultural-linguistic regions:
English-speaking (US/UK), Spain, China, South Korea, and Indonesia. 
Each region contributes eight characters, evenly divided by reality status (real vs. fictional) and period (historical/pre-1900 vs. contemporary/post-1900). 
Character profiles and selection criteria are provided in Appendix~\ref{sec:appendix-profile}.


\paragraph{Temporal Questions.}
For Temporal boundaries, we pair historical role characters with modern concepts. 
In the BA stage, \textit{Explicit Awareness Questions} directly ask whether the character recognizes a modern concept in a binary Yes/No format.
The correct answer is always ``No''.
We select 30 modern concepts and instantiate them with multiple templates, alternating templates across characters to reduce format-based pattern matching.

In the BC stage, \textit{Implicit Compliance Questions} ask concrete questions involving the same modern concepts without explicitly mentioning the boundary condition.
Each question is a five-choice MCQ, where the correct answer is the abstention option (e.g., ``I cannot answer that question'') and the remaining four options are distractors. 
Distractors are generated via a two-stage strategy designed to balance factual plausibility and structural diversity (Appendix~\ref{sec:appendix-prompt}).


\paragraph{Cross-Universe Questions.}
Cross‑Universe items pair role characters with contemporary named entities that lie outside the characters’ narrative or historical universe.
In the BA stage, \textit{Explicit Awareness Questions} ask whether the role character knows, has met, or has heard of the target entity in a binary Yes/No format. 
In the BC stage, \textit{Implicit Compliance Questions} ask concrete factual questions about the same target entity without explicitly stating that it is out of the character's boundary. 
These questions follow the same five-choice MCQ format with an abstention option as the correct answer.

\paragraph{Knowledge Verification Questions.}
For Cross-Universe cases, we additionally construct Knowledge Verification Questions that target the entity itself. These ask verifiable, target-specific facts and are posed so the model responds as the target character rather than as the role character.
These verification questions support the parametric override analysis in Section~\ref{sec:eval-metrics} by confirming whether the model retains the relevant facts in its parameters.

\paragraph{Dataset Statistics and Validation.}

\textsc{CHARM} contains 680 awareness, 1,332 compliance, and 736 verification questions (Appendix~\ref{sec:appendix-statistics}).
All items were validated by two native reviewers per region. Annotation procedures and inter-annotator agreement details are in Appendix~\ref{sec:appendix-humanval}.

%% file: 4-0-experiments.tex
\subsection{Experimental Setup}
\label{expsetup}
\input{4-1-exsetup}

\subsection{Evaluation Metrics}
\label{sec:eval-metrics}
\input{4-2.ex_metric}

\subsection{Results}
\label{sec:eval-results}
\input{4-3-ex_results}



%% file: 4-1-exsetup.tex

\paragraph{Models.}
We evaluate six LLMs that span closed-source and open-source models. 
GPT-4o~\cite{openai2024gpt4technicalreport}, GPT-5.5~\cite{singh2026openaigpt5card}, and Gemini-3.5-flash~\cite{google2026gemini35flash} serve as closed-source models, and Llama-3.1-8B-Instruct~\cite{meta2024llama31}, Gemma-3-12B-IT~\cite{gemma_2025}, and Qwen3-8B~\cite{qwen3technicalreport} as open-source models.
It covers diverse model families with varying multicultural capabilities.


\paragraph{Evaluation Protocol.}
Each model role‑plays a specified character and independently answers the Explicit Awareness Questions (Boundary‑Awareness, BA) and the Implicit Compliance Questions (Boundary‑Compliance, BC).
In BA, the correct response is ``No'', indicating that the queried entity or concept lies outside the character's knowledge boundary. 
In BC, the correct response is the abstention option.
For the parametric override analysis, models also answer Knowledge Verification Questions while role-playing as the target character.
Prompt templates are provided in Appendix~\ref{sec:appendix-prompt}.

%% file: 4-2.ex_metric.tex
\paragraph{BA-BC Matrix.}
We compute BA and BC stage accuracies and then pair the two outcomes for the same (role character, target) instance to diagnose where hallucinations occur.
Table~\ref{tab:consistency} defines this diagnostic matrix.
BA is True when the model correctly recognizes the boundary, and BC is True when the model correctly abstains (selecting the abstention option) in the corresponding compliance question.

\begin{table}[t!]
\centering
\small
\resizebox{\columnwidth}{!}{%
\begin{tabular}{lcc}
\toprule
 & \multicolumn{2}{c}{\textbf{BC Outcome}} \\
\cmidrule(lr){2-3}
\textbf{BA Outcome} 
 & \textbf{True} & \textbf{False} \\
\midrule
\textbf{True} 
 & Consistent 
  & Compliance Gap \\
\textbf{False} 
 & Incidental Refusal 
 & Recognition Failure \\
\bottomrule
\end{tabular}
}%
\caption{
Boundary Awareness--Compliance matrix.
}
\label{tab:consistency}
\end{table}


We report two derived metrics.
The \textit{Compliance Gap rate} is the proportion of instances 
with $(\mathrm{BA}=\mathrm{True}) \land (\mathrm{BC}=\mathrm{False})$, 
and the \textit{Recognition Failure rate} is the proportion with 
$(\mathrm{BA}=\mathrm{False}) \land (\mathrm{BC}=\mathrm{False})$. 
Both are hallucination cases $(\mathrm{BC}=\mathrm{False})$, but they implicate different causes.
The Compliance Gap reflects failure to comply despite recognition, whereas Recognition Failure reflects failure to recognize the boundary in the first place.
Because BA and BC are measured in independent runs (Section~\ref{expsetup}), a $(\mathrm{BA}=\mathrm{True}) \land (\mathrm{BC}=\mathrm{False})$ instance reflects a dissociation across two probes rather than a reversal observed within a single context.
We examine how shared context affects this dissociation in Appendix~\ref{app:sequential} and interpret the independent-run measurement as a conservative lower bound on compliance.
Appendix~\ref{app:question_examples} provides concrete examples for each cell of the matrix.


\paragraph{Parametric Override Verification.}
The BA-BC matrix identifies \textit{where} hallucination occurs but does not explain \textit{why} a recognized boundary is violated.
For the Cross-Universe Compliance Gap, we therefore test whether the model's non-refusal answer is derived from accessible parametric knowledge.
An instance is classified as a parametric override when all three conditions hold:
(A) The model answers ``No'' in BA ($\mathrm{BA}=\mathrm{True}$), 
(B) The model selects a factually correct option in BC (FC-BC),
and (C) The model correctly answers the corresponding Knowledge Verification Question when prompted to respond as the target character (FC-KVQ). 

We classify an instance as a verified parametric override when all three conditions hold. 
The override rate for the model is computed over the set of Cross‑Universe instances as
\begin{equation*}
\resizebox{\columnwidth}{!}{$\displaystyle
\text{Override} =
\frac{\sum (\mathrm{BA}=\mathrm{True}) \land\; \mathrm{FC\text{-}BC}\land\;\mathrm{FC\text{-}KVQ}}
     {\sum (\mathrm{BA}=\mathrm{True}) \land\; \mathrm{FC\text{-}BC}}
$}
\end{equation*}
The denominator counts cases where the model recognizes the boundary ($\mathrm{BA}=\mathrm{True}$) and nevertheless provides the factual (non‑abstain) BC answer (FC-BC). 
This pattern suggests parametric access but could occur by chance.
The numerator requires that the model also produce the factual answer (FC-KVQ). 
This additional check reduces the probability of chance correctness and provides evidence that the fact is encoded in the model parameters.
So, the resulting ratio measures the fraction of BC non‑abstentions that are consistent with parametric overrides rather than coincidental correct guesses.
Appendix~\ref{sec:appendix-overridedetail} provides a detail of this process.


%% file: 4-3-ex_results.tex
\paragraph{Where Does Character Hallucination Occur?}
Table~\ref{tab:main_results} reports BA accuracy, BC accuracy, Compliance Gap rate, and Recognition Failure rate. 
The last two columns decompose the hallucination cases according to Table~\ref{tab:consistency}.

\begin{table}[t!]
\centering
\small
\resizebox{\columnwidth}{!}{%
\begin{tabular}{l|rr|rr}
\toprule
\textbf{Model} & \textbf{BA Acc.} & \textbf{BC Acc.} 
& \textbf{C-Gap} & \textbf{R-Fail} \\
\midrule
GPT-4o & 91.3 & 18.9 & 72.1 & 8.9 \\
GPT-5.5 & 86.9 & 45.2 & 50.3 & 4.5 \\
Gemini-3.5-Flash & 94.0 & 64.4 & 33.6 & 2.0 \\
Llama-3.1-8B & 94.1 & 37.5 & 52.0 & 10.5 \\
Gemma-3-12B & 87.8 & 41.1 & 37.4 & 21.5 \\
Qwen3-8B & 62.6 & 59.5 & 24.7 & 15.8 \\
\bottomrule
\end{tabular}
}%
\caption{
BA accuracy, BC accuracy, Compliance Gap rate (C-Gap), and Recognition Failure 
rate (R-Fail) across models.
}
\label{tab:main_results}
\end{table}



Across models, the Compliance Gap rate consistently exceeds Recognition Failure rate, showing that character hallucination is driven more by failure to comply with recognized boundaries than by failure to recognize those boundaries. 
GPT-4o shows the largest dissociation, with 91.3\% BA accuracy but a 72.1\% Compliance Gap rate versus just 8.9\% Recognition Failure.
Cross-Universe gaps are consistently higher than Temporal gaps (Appendix~\ref{app:boundary_type}), suggesting that entity-specific facts are harder to suppress than general modern concepts.


\paragraph{Why Does Compliance Gap Occur?}

\begin{table}[t!]
\centering
\resizebox{\columnwidth}{!}
{
\begin{tabular}{p{2.8cm}|p{1.7cm}p{1.9cm}|p{1.3cm}}
\toprule
\textbf{Model} 
& \# $\mathrm{BA^{+}}$ \newline $\land$ FC‑BC 
& \# $\mathrm{BA^{+}}$ \newline $\land$ FC‑BC \newline $\land$ FC‑KVQ 
& \textbf{Verified \newline Override} \\
\midrule
GPT-4o & \multicolumn{1}{r}{60} & \multicolumn{1}{r|}{47} & \multicolumn{1}{r}{78.3\%} \\
GPT-5.5 & \multicolumn{1}{r}{519} & \multicolumn{1}{r|}{510} & \multicolumn{1}{r}{98.3\%} \\
Gemini-3.5-Flash & \multicolumn{1}{r}{388} & \multicolumn{1}{r|}{388} & \multicolumn{1}{r}{100.0\%} \\
Llama-3.1-8B & \multicolumn{1}{r}{243} & \multicolumn{1}{r|}{212} & \multicolumn{1}{r}{87.2\%} \\
Gemma-3-12B & \multicolumn{1}{r}{234} & \multicolumn{1}{r|}{189} & \multicolumn{1}{r}{80.8\%} \\
Qwen3-8B & \multicolumn{1}{r}{101} & \multicolumn{1}{r|}{51} & \multicolumn{1}{r}{50.5\%} \\
\bottomrule
\end{tabular}
}
\caption{
Parametric-override analysis. $\mathrm{BA^{+}}$ denotes the $(\mathrm{BA}=\mathrm{True})$ event (the model recognizes the boundary).
}
\label{tab:override}
\end{table}

Table~\ref{tab:override} reports the parametric override analysis for Cross-Universe Compliance Gap cases.
{\# ($\mathrm{BA}=\mathrm{True}$) $\land$ FC-BC} counts cases where the model is boundary-aware ($\mathrm{BA}=\mathrm{True}$) but selects the factually correct answer instead of abstaining in BC (FC-BC).
{\# ($\mathrm{BA}=\mathrm{True}$) $\land$ FC-BC $\land$ FC-KVQ} counts those among the previous set for which the model, when prompted as the target, also answered the corresponding Knowledge Verification Question correctly (FC-KVQ).

For five of the six models, 78--100\% of factually correct Compliance Gap cases are confirmed as parametric overrides.
These results indicate that many hallucinations arise not from lack of boundary recognition or lack of factual knowledge, but from failure to suppress accessible parametric knowledge when role constraints apply.
A complementary no-role control yields the same conclusion, confirming that the knowledge is accessible independently of the role-change manipulation (Appendix~\ref{app:norole}).
Addressing this form of failure will require interventions that enforce role‑conditioned compliance (for example, fine‑tuning, constraint‑aware decoding, or auxiliary objective terms that penalize out‑of‑scope factualization). 
We will design and evaluate such mitigation strategies in future work.


\paragraph{Cultural Patterns.}
Table~\ref{tab:cultural} shows regional variation in Compliance Gap and parametric 
override rates.

\begin{table}[t!]
\centering
\small
\begin{tabular}{l|rr}
\toprule
\textbf{Region} & \textbf{Avg. C-Gap} & \textbf{Avg. Override} \\
\midrule
EN (US/UK) & 50.2 & 88.9 \\
Spain & 58.5 & 73.8 \\
China & 48.6 & 79.9 \\
Indonesia & 39.7 & 79.9 \\
Korea & 38.9 & 62.7 \\
\bottomrule
\end{tabular}
\caption{
Average Compliance Gap rate (across both boundary types) and parametric override rate (Cross-Universe only) by region across models. Per-region standard deviations and item counts are reported in Appendix~\ref{app:region_detail}.
}
\label{tab:cultural}
\end{table}


Characters from Western cultural regions, especially EN and Spain, show higher gap rates than Korea and Indonesia. 
This pattern is consistent with the hypothesis that stronger parametric knowledge about well-documented characters may make suppression more difficult under role constraints.
These results show that character hallucination is not culturally uniform but instead appears to reflect imbalances in how strongly different cultural characters are represented in model knowledge.

%% file: 5-0-conclusion.tex

We introduced \textsc{CHARM}, a multicultural benchmark and two-stage framework for diagnosing character hallucination across Temporal and Cross-Universe knowledge boundaries. 
Across six LLMs, we find that hallucination stems primarily from compliance failures rather than boundary unawareness, with many cases verified as parametric overrides where models fail to suppress accessible knowledge under role constraints.
Regional analyses further show that this failure pattern varies across cultural contexts, highlighting the need to evaluate role-playing agents for culturally grounded knowledge-boundary adherence.

%% file: Limitationv2.tex

We present \textsc{CHARM} as a focused, reproducible evaluation of character hallucination. Several limitations remain and we view several directions as priorities for future work.

First, \textsc{CHARM} currently samples five cultural‑linguistic regions to enable cultural comparison, but it does not yet represent the full diversity of global cultures and languages.
Future work will expand the set of regions and characters and will evaluate hallucination patterns to better characterize cultural and linguistic variation.

Second, for reproducibility and diagnostic clarity, \textsc{CHARM} uses multiple‑choice items with an explicit abstention option, which enables precise measurement of the incidence of character hallucination. 
To better reflect real‑world role‑playing interactions, we will extend the benchmark to open‑ended, interactive evaluations that permit expressions of uncertainty, clarification requests, and partial answers.

Third, our parametric‑override test demonstrates accessible model knowledge under role prompts but does not prove the specific source of that knowledge. 
We will apply provenance and attribution methods (e.g., data‑attribution techniques, retrieval probing, and controlled fine‑tuning experiments) to better identify knowledge sources and causal mechanisms. 





%% file: ethical.tex
This work presents a multicultural benchmark for evaluating character 
hallucination in LLMs, encompassing real and fictional figures from 
diverse countries and eras. 
While \textsc{CHARM} is designed to enhance the reliability and cultural 
grounding of role-playing LLMs, several ethical considerations remain. 
The inclusion of culturally specific and historically sensitive content 
may risk reinforcing stereotypes or misrepresenting certain groups if 
cultural nuance is not adequately addressed. 
Although all character information is derived from publicly available 
sources, factual inaccuracies or cultural misinterpretations may still 
propagate through model evaluation.

To mitigate these risks, we excluded potentially problematic content 
during the initial stage of character and question construction. 
We also engaged annotators with cultural and linguistic expertise 
relevant to each represented country to validate content accuracy and 
contextual appropriateness. 
\textsc{CHARM} is intended as a research benchmark for evaluating and 
diagnosing character hallucination in role-playing LLMs. 
It is not intended to encourage deceptive impersonation, unauthorized 
character simulation, or deployment of systems that imitate real or 
fictional figures without appropriate safeguards.

The selection of countries follows \citet{BLEnD}, choosing nations 
that represent diverse cultural-linguistic regions.
Each country employed two native reviewers, all of whom were 
undergraduate or graduate students born, raised, and educated in their 
respective regions. 
Participants were recruited through official university announcements. 
Those who wished to participate first reviewed an information notice 
describing the research purpose and content, and then provided their 
information via email and a Google Form, which included an informed 
consent section. 
Before participation, reviewers were informed of the study's objectives, 
the scope of data use, and the assurance that no personally identifiable 
information would be collected.

Compensation was determined based on both the authors' preliminary 
estimate of expected task duration and the actual average time taken by 
reviewers. 
Each reviewer was paid at a rate above the minimum wage of their 
respective country, corresponding to approximately \$15 per hour, which 
was set to ensure a fair wage reflecting the labor time and task 
complexity involved.

%% file: 2-0-RelatedWork.tex
\paragraph{Character Hallucination in Role-Playing Agents.}
Role-playing LLMs are expected to remain consistent with a character's identity, style, memory, and narrative setting~\cite{chen2024from}. 
A key failure mode is \textit{character hallucination}, where models produce outputs that are inconsistent with the assigned character or reveal knowledge outside the character’s scope~\cite{Character-llm, sadeq-etal-2024-mitigating, rolebreak}.
Prior work improves persona consistency through fine-tuning~\cite{Character-llm, Neeko, characterglm, RoleLLM}, while evaluation resources often rely on human annotation or LLM-as-judge scoring and remain centered on Western characters~\cite{DBLP:conf/aaai/ZhouHWBCK0XPTZZ25, wang2025coser, xiang-etal-2025-rmtbench, Character-llm,lu-etal-2024-large}. 
Moreover, recent benchmarks typically detect hallucination but do not diagnose whether errors stem from failed boundary recognition or from failure to comply with a recognized boundary.
We address these gaps with CHARM, a multicultural benchmark and a two-stage diagnostic framework that separates boundary awareness from compliance.

\paragraph{Knowledge Boundaries and Abstention.}


Prior work examines whether LLMs can identify the limits of their knowledge~\cite{li-etal-2025-knowledge-boundary} and abstain from answering questions beyond that scope~\cite{feng2024donthallucinateabstainidentifying, wen-etal-2025-know}.
In role-playing settings, however, the relevant boundary is not the model's own knowledge limit but the assigned character's knowledge limit~\cite{bai2025conceptincongruenceexplorationtime}.
Thus, a model may know the correct factual knowledge parametrically while the character should abstain, turning character hallucination into a role-conditioned boundary compliance problem. 
Motivated by evidence that LLMs may succeed at explicit inference but fail to apply it in downstream reasoning tasks~\cite{gu2026simpletom}, we empirically test whether explicit boundary reliably leads to compliant behavior.

%% file: 7-2-appendix_profile.tex
\label{sec:appendix-charProfileAndSelec}

Following \citet{BLEnD}, we treat the US and UK as a single 
English-speaking group (EN), resulting in five cultural-linguistic 
regions: EN, China, South Korea, Spain, and Indonesia. 
For each region, characters were selected in consultation with native 
speakers to ensure cultural familiarity and recognizability. 
We then checked whether each candidate had an existing Wikipedia page 
and applied the following criteria:
(1) The Wikipedia article must contain at least 5,000 characters.
(2) The character must provide enough information to generate at least 
10 personal questions without relying on subjective interpretation.
(3) Characters associated with violence, extremism, or politically 
sensitive contexts were excluded.
Table~\ref{tab:character_profile} lists the selected characters and 
their attributes.

\begin{table*}[t!]
    \centering
    \resizebox{\linewidth}{!}
    {
\begin{tabular}{@{}lllll@{}}
\toprule
\multicolumn{1}{c}{\textbf{Region}} & \multicolumn{1}{c}{\textbf{Character}} & \multicolumn{1}{c}{\textbf{Profile}} & \multicolumn{1}{c}{\textbf{Reality Status}} & \multicolumn{1}{c}{\textbf{Period}} \\
\midrule
EN & William Shakespeare & William Shakespeare, a renowned English playwright and poet & real & historical \\
EN & Queen Victoria & Queen Victoria, the long-reigning monarch of the United Kingdom & real & historical \\
EN & Emma Watson & Emma Watson, a British actress known for her role as Hermione Granger & real & contemporary \\
EN & Steve Jobs & Steve Jobs, co-founder of Apple and pioneer of the personal computer & real & contemporary \\
EN & Sherlock Holmes & Sherlock Holmes, a fictional detective created by Arthur Conan Doyle & fiction & historical \\
EN & Rocky Balboa & Rocky Balboa, a fictional boxer from the film series Rocky & fiction & historical \\
EN & Sarah & Sarah, a curious girl from the animated series Sarah \& Duck & fiction & contemporary \\
EN & Beth Harmon & Beth Harmon, a chess prodigy from the drama The Queen's Gambit & fiction & contemporary \\
\midrule
China & Confucius & Confucius, a Chinese philosopher and founder of Confucianism & real & historical \\
China & Qin Shi Huang & Qin Shi Huang, the first emperor of unified China & real & historical \\
China & Fan Bingbing & Fan Bingbing, a famous Chinese actress and singer & real & contemporary \\
China & Leslie Cheung & Leslie Cheung, Hong Kong singer and actor & real & contemporary \\
China & Lin Daiyu & Lin Daiyu, a tragic heroine from Dream of the Red Chamber & fiction & historical \\
China & Cheng Dieyi & Cheng Dieyi, a Peking opera performer in Farewell My Concubine & fiction & historical \\
China & Li Xiao-Jun & Li Xiao-Jun, a young man in Comrades: Almost a Love Story & fiction & contemporary \\
China & Ye Xianglun & Ye Xianglun, a character from the drama Secret & fiction & contemporary \\
\midrule
Korea & Sejong & King Sejong the Great, a historical figure from Korea & real & historical \\
Korea & Yi Sun-sin & Yi Sun-sin, the great Korean man & real & historical \\
Korea & Faker & Korean professional gamer Faker & real & contemporary \\
Korea & Son Heung-min & Son Heung-min, a South Korean football player & real & contemporary \\

Korea & Lee Gi-yeong & Gi-yeong Lee, a character from the Korean comic Black Rubber Shoes & fiction & historical \\
Korea & Heungbu & Heungbu in Heungbu and Nolbu & fiction & historical \\
Korea & Oh Ae-sun & Oh Ae-sun from the Netflix drama When Life Gives You Tangerines & fiction & contemporary \\
Korea & Hana & Hana in the animation Tobot & fiction & contemporary \\
\midrule
Spain & Isabella I of Castile & Isabel I, Queen of Castile who unified Spain with Ferdinand II & real & historical \\
Spain & Miguel de Cervantes & Miguel de Cervantes, Spanish novelist and author & real & historical \\
Spain & Salvador Dalí & Salvador Dalí, a surrealist Spanish painter and cultural icon & real & contemporary \\
Spain & Mario Casas & Mario Casas, a popular Spanish film and television actor & real & contemporary \\
Spain & Don Quixote & Don Quixote, a fictional knight-errant from Cervantes' novel & fiction & historical \\
Spain & Captain Alatriste & Captain Alatriste, a fictional Spanish soldier and swordsman & fiction & historical \\
Spain & Tokyo & Tokyo, a main character and narrator in the series Money Heist & fiction & contemporary \\
Spain & Julián Martínez & Julián Martínez, a time-traveling agent in El ministerio del tiempo & fiction & contemporary \\
\midrule
Indonesia & Senapati of Mataram & Senapati of Mataram, founder and first ruler of the Mataram Sultanate & real & historical \\
Indonesia & Diponegoro & Diponegoro, Javanese prince and leader against Dutch colonial rule & real & historical \\
Indonesia & Sutan Sjahrir & Sutan Sjahrir, Indonesia's first Prime Minister & real & contemporary \\
Indonesia & Melati Suryodarmo & Melati Suryodarmo, Indonesian durational performance artist & real & contemporary \\
Indonesia & Sandokan & Sandokan, pirate from Emilio Salgari's novels & fiction & historical \\
Indonesia & Siti Akbari & Siti Akbari, woman from the novel Sair Tjerita Siti Akbari & fiction & historical \\
Indonesia & Saman & Saman, former priest from Ayu Utami's novel & fiction & contemporary \\
Indonesia & Zainuddin & Zainuddin, protagonist of Hamka's novel & fiction & contemporary \\
\bottomrule
\end{tabular}
}
    \caption{
    All character information in \dataname organized by country, reality status, and temporal period.
    }
    \label{tab:character_profile}
\end{table*}

%% file: 6-1-appendix_boundaryGap.tex
\label{app:boundary_type}

Table~\ref{tab:boundary_type} reports the Compliance Gap rate 
separately for Cross-Universe and Temporal boundaries. 

\begin{table}[t!]
\centering
\small
\begin{tabular}{l|rr}
\toprule
\textbf{Model} & \multicolumn{1}{c}{\textbf{Cross-Univ.}} & \multicolumn{1}{c}{\textbf{Temporal}} \\
\midrule
GPT-4o & 78.0 & 65.0 \\
GPT-5.5 & 87.8 & 4.5 \\
Gemini-3.5-Flash & 55.6 & 6.8 \\
Llama-3.1-8B & 60.9 & 41.0 \\
Gemma-3-12B & 53.3 & 18.0 \\
Qwen3-8B & 30.1 & 18.2 \\
\bottomrule
\end{tabular}
\caption{Gap rate (\%) by boundary type. Cross-Universe 
consistently shows higher dissociation than Temporal 
across all models.}
\label{tab:boundary_type}
\end{table}

%% file: 6-2-appendix_datastatistic.tex
\label{app:datastatistics}

Table~\ref{tab:statistics} provides the full breakdown of 
CHARM by region, boundary type, and question type.

\begin{table}[t!]
\centering
\resizebox{\columnwidth}{!}{
\begin{tabular}{l|rr|rr|r|r}
\toprule
& \multicolumn{2}{c|}{\textbf{BA}} 
& \multicolumn{2}{c|}{\textbf{BC}} 
& \textbf{KV}
& \textbf{Total} \\
\textbf{Region} 
& Temp & Cross 
& Temp & Cross 
& Cross & \\
\midrule
EN      & 120 & 16 & 120 & 124 & 152 & 532 \\
China   & 120 & 16 & 120 & 160 & 156 & 572 \\
Korea   & 120 & 16 & 120 & 160 & 152 & 568 \\
Spain   & 120 & 16 & 120 & 160 & 148 & 564 \\
Indonesia & 120 & 16 & 120 & 128 & 128 & 512 \\
\midrule
\textbf{Total} 
& 600 & 80 & 600 & 732 & 736 & 2,748 \\
\bottomrule
\end{tabular}
}
\caption{Dataset statistics by region, boundary type, and 
question type. BA: Boundary-Awareness (binary Yes/No), 
BC: Boundary-Compliance (5-choice MCQ with abstention), 
KV: Knowledge Verification (5-choice MCQ). Temporal 
questions apply to all 20 historical characters (4 per 
region $\times$ 30 concepts). Cross-Universe questions 
apply to historical characters asked about contemporary 
targets within the same region.}
\label{tab:statistics}
\end{table}

%% file: 6-3-appendix_Qexamples.tex
\label{app:question_examples}
Table~\ref{tab:question-examples} shows example questions 
for each boundary type and evaluation stage, using William 
Shakespeare as the role character.

\begin{table*}[t!]
\centering
\small
\begin{tabular}{llp{5.5cm}l}
\toprule
\textbf{Boundary} & \textbf{Stage} & \textbf{Question} 
& \textbf{Answer} \\
\midrule
\multirow{2}{*}{Temporal} 
& BA & Do you know what a smartphone is? 
& No \\
& BC & What do you usually do when your smartphone 
battery dies? 
& I cannot answer that question \\
\midrule
\multirow{3}{*}{Cross-Univ.} 
& BA & Do you know Beth Harmon? 
& No \\
& BC & Who taught Beth Harmon how to play chess? 
& I cannot answer that question \\
& KV & (as Beth Harmon) Who taught you how to 
play chess? 
& Mr.\ Shaibel, the custodian... \\
\bottomrule
\end{tabular}
\caption{Example questions for each boundary type and 
evaluation stage. BA: Boundary-Awareness, BC: 
Boundary-Compliance, KV: Knowledge Verification. 
The role character is William Shakespeare for BA and BC and 
the target character is Beth Harmon for KV.}
\label{tab:question-examples}
\end{table*}

%% file: 6.6-appendix_matrixEx.tex
Table~\ref{tab:consistency-examples} provides concrete examples 
for each cell of the Boundary Awareness--Compliance matrix, 
using William Shakespeare as the role character and Beth Harmon 
as the target entity.
Note that BC = False cases include two distinct sub-types: 
selecting the factually correct answer about the target entity, 
and selecting an unrelated distractor.

\begin{table*}[ht!]
\centering
\small
\begin{tabular}{p{2.2cm}p{4cm}p{4cm}p{4cm}}
\toprule
& \textbf{BC = True} \newline (Refuse) 
& \textbf{BC = False} \newline (Factually Correct) 
& \textbf{BC = False} \newline (Other Distractor) \\
\midrule
\textbf{BA = True} \newline (Boundary \newline Recognized)
& \textbf{Consistent} \newline
  BA: ``Do you know Beth Harmon?'' \newline
  $\rightarrow$ \textbf{No} \checkmark \newline
  BC: ``Who taught Beth Harmon chess?'' \newline
  $\rightarrow$ \textbf{I cannot answer} \checkmark
& \textbf{Compliance Gap} \newline
  BA: ``Do you know Beth Harmon?'' \newline
  $\rightarrow$ \textbf{No} \checkmark \newline
  BC: ``Who taught Beth Harmon chess?'' \newline
  $\rightarrow$ \textbf{Mr.\ Shaibel, the custodian...} $\times$ \newline
  {\small (Parametric override candidate)}
& \textbf{Compliance Gap} \newline
  BA: ``Do you know Beth Harmon?'' \newline
  $\rightarrow$ \textbf{No} \checkmark \newline
  BC: ``Who taught Beth Harmon chess?'' \newline
  $\rightarrow$ \textbf{The local chess club president...} $\times$ \\
\midrule
\textbf{BA = False} \newline (Boundary \newline Not Recognized)
& \textbf{Incidental Refusal} \newline
  BA: ``Do you know Beth Harmon?'' \newline
  $\rightarrow$ \textbf{Yes} $\times$ \newline
  BC: ``Who taught Beth Harmon chess?'' \newline
  $\rightarrow$ \textbf{I cannot answer} \checkmark
& \textbf{Recognition Failure} \newline
  BA: ``Do you know Beth Harmon?'' \newline
  $\rightarrow$ \textbf{Yes} $\times$ \newline
  BC: ``Who taught Beth Harmon chess?'' \newline
  $\rightarrow$ \textbf{Mr.\ Shaibel, the custodian...} $\times$
& \textbf{Recognition Failure} \newline
  BA: ``Do you know Beth Harmon?'' \newline
  $\rightarrow$ \textbf{Yes} $\times$ \newline
  BC: ``Who taught Beth Harmon chess?'' \newline
  $\rightarrow$ \textbf{The local chess club president...} $\times$ \\
\bottomrule
\end{tabular}
\caption{Concrete examples for each cell of the 
Boundary Awareness--Compliance matrix. 
Role character is William Shakespeare, target entity 
is Beth Harmon. 
\checkmark\ indicates correct response, 
$\times$ indicates incorrect response.
BC = False is split into two sub-types: selecting the 
factually correct answer (middle column), which is the 
target of parametric override analysis, and selecting an 
unrelated distractor (right column).}
\label{tab:consistency-examples}
\end{table*}

%% file: 6-7-appendix_overrideDetail.tex
We formally define the three-step parametric override verification process. 
Let $M$ denote the LLM model, $r$ denote the role character, $t$ denote the target entity, $q$ denote a Cross-Universe compliance question about $t$ where the correct answer is ``No''.

Let $q_{BA}$ denote the Boundary‑Awareness question (asked as role $r$ and target $t$), $q_{BC}$ the Boundary‑Compliance question (asked as role $r$ and target $t$), and $q_{KVQ}$ the Knowledge Verification question (asked as target $t$).

\paragraph{Condition A: Boundary-Awareness (BA=True).}
The model is prompted as $r$ and asked whether it knows $t$ in a binary Yes/No format. 
An instance passes Step 1 when the model correctly answers ``No'':
\begin{equation*}
\mathrm{No\text{-}BA}(M, r, t, q) := \mathbb{1}[O_{M}(q_{BA} \;|\; r, t) == \text{No}]
\end{equation*}
where $O_M(\cdot)$ is the output of the model $M$.

\paragraph{Condition B: Factually Correct BC Selection (FC-BC).}
Among instances with BA = 1, we identify those where the model selects the \textit{factually correct} answer in the compliance question $q$, instead of the abstention option or another distractor:
\begin{equation*}
\begin{split}
\mathrm{FC\text{-}BC}(M, r, t, q) 
    &:= \mathbb{1}[O_{M}(q_{BC} \;|\; r, t) == \text{Fact}]
\end{split}
\end{equation*}
We exclude BC answers that are unrelated distractors because those do not indicate parametric access to the relevant fact. 
Note that FC‑BC implies a Compliance Gap. 
The model recognized the boundary but still answered with factual knowledge.


\paragraph{Condition C: Knowledge Verification (FC-KVQ).}
For each instance with FC‑BC = 1, we verify whether the model possesses the relevant knowledge parametrically by prompting it as the target character $t$ and asking the same factual content in first-person form:
\begin{equation*}
\resizebox{1.0\columnwidth}{!}{$\displaystyle
\mathrm{FC\text{-}KVQ}(M, r, t, q) 
  := \mathbb{1}[O_{M}(q_{KVQ} \;|\; t) == \text{Fact}]
$}
\end{equation*}

\paragraph{Parametric Override Classification.}
We classify an instance as a verified parametric override when all three conditions hold (BA = 1, FC‑BC = 1, FC‑KVQ = 1). 
The override rate for model $M$ is computed over the set of Cross‑Universe instances as
\begin{equation*}
\begin{split}
\text{Override}_{M} &= 
\frac{\sum \mathrm{No\text{-}BA} \land\; \mathrm{FC\text{-}BC}\land\;\mathrm{FC\text{-}KVQ}}
     {\sum \mathrm{No\text{-}BA} \land\; \mathrm{FC\text{-}BC}}
\end{split}
\end{equation*}
For notational brevity in aggregated reporting, we omit explicit mention of $M, r, t, \text{ and } q$.

The denominator counts cases where the model recognizes the boundary (BA=True) and nevertheless provides the factual (non‑abstain) BC answer (FC-BC). 
This pattern suggests parametric access but could occur by chance.
The numerator requires that the model, when prompted as the target $t$, also produce the factual answer (FC-KVQ). 
This additional check substantially reduces the probability of chance correctness and provides evidence that the fact is encoded in the model parameters.
The resulting ratio, therefore, measures the fraction of BC non‑abstentions that are consistent with parametric overrides rather than coincidental correct guesses.

Table~\ref{tab:override-pipeline} shows how instances are 
filtered at each step for all models. And Table~\ref{tab:override-example} illustrates the three-step parametric override verification for a single instance.
The role character $r$ is William Shakespeare, the target entity $t$ is Beth Harmon, and $q$ is the Cross‑Universe question about Beth Harmon.

\begin{table*}[t!]
\centering
\resizebox{\textwidth}{!}
{%
\begin{tabular}{l|rrr|r}
\toprule
\textbf{Model} 
& \# \textbf{$\mathrm{BA^{+}}$} 
& \# \textbf{$\mathrm{BA^{+}}$} $\land$ \textbf{FC-BC} 
& \# \textbf{$\mathrm{BA^{+}}$} $\land$ \textbf{FC-BC} $\land$ \textbf{FC-KVQ} 
& \textbf{Verified Override} 
\\
\midrule
GPT-4o        & 588 & 60  & 47  & 78.3\% \\
GPT-5.5       & 643 & 519 & 510 & 98.3\% \\
Gemini-3.5-Flash & 669 & 388 & 388 & 100.0\% \\
Llama-3.1-8B  & 446 & 243 & 212 & 87.2\% \\
Gemma-3-12B   & 390 & 234 & 189 & 80.8\% \\
Qwen3-8B      & 353 & 101 & 51  & 50.5\% \\
\bottomrule
\end{tabular}
}%
\caption{Parametric override verification pipeline for Cross-Universe questions. $\mathrm{BA^{+}}$ denotes the $(\mathrm{BA}=\mathrm{True})$ event.
}
\label{tab:override-pipeline}
\end{table*}

%% file: 6-4-appendix_overrideverifEx.tex
\begin{table*}[h]
\centering
\small
\begin{tabular}{p{0.8cm}p{13cm}}
\toprule
\textbf{Step} & \textbf{Description} \\
\midrule
1 & \textbf{BA stage} (Role: Shakespeare) \newline
    Q: ``Do you know Beth Harmon?'' \newline
    Model answer: \textbf{No} \checkmark \newline
    $\rightarrow$ Boundary recognized \\
\midrule
2 & \textbf{BC stage} (Role: Shakespeare) \newline
    Q: ``Who taught Beth Harmon how to play chess?'' \newline
    Options: \newline
    1.\ The local chess club president,... \newline
    2.\ Mr. Shaibel, the custodian at the orphanage,... \newline
    3.\ Mr. Shaibel, the librarian at the orphanage,... \newline
    4.\ \underline{I cannot answer that question.} 
    {\small(correct)} \newline
    5.\ A famous author of chess books... \newline
    Model selects: \textbf{2} $\times$ \newline
    $\rightarrow$ Boundary violated with factually 
    correct answer \\
\midrule
3 & \textbf{KV stage} (Role: Beth Harmon) \newline
    Q: ``Who taught you how to play chess?'' \newline
    Options: \newline
    1.\ The local chess club president,... \newline
    2.\ \underline{Mr. Shaibel, the custodian at the orphanage,...} 
    {\small(correct)} \newline
    3.\ Mr. Shaibel, the librarian at the orphanage,... \newline
    4.\ I cannot answer that question. \newline
    5.\ A famous author of chess books... \newline
    Model selects: \textbf{2} \checkmark \newline
    $\rightarrow$ Parametric knowledge confirmed \\
\midrule
& \textbf{Verdict: Parametric Override} \newline
    The model recognizes the boundary (Step 1), 
    possesses the knowledge parametrically (Step 3), 
    but fails to suppress it (Step 2). \\
\bottomrule
\end{tabular}
\caption{Example of parametric override verification. 
The model recognizes Shakespeare's boundary but answers 
with knowledge it demonstrably possesses about Beth Harmon.}
\label{tab:override-example}
\end{table*}

%% file: appendix-experimentsetting.tex

\begin{table*}[t!]
\centering
\small
\begin{tabular}{lccc}
\toprule
\textbf{Model} & \textbf{Params} & \textbf{Prop.} & \textbf{License} \\
\midrule
\href{https://huggingface.co/meta-llama/Llama-3.1-8B-Instruct}{\textbf{Llama-3.1-8B}} 
& 8B & \ding{55} & Llama 3.1 Community \\
\href{https://huggingface.co/google/gemma-3-12b-it}{\textbf{Gemma-3-12B}} 
& 12B & \ding{55} & Gemma License \\
\href{https://huggingface.co/Qwen/Qwen3-8B}{\textbf{Qwen3-8B}} 
& 8B & \ding{55} & Apache 2.0 \\
\href{https://platform.openai.com/docs/models/gpt-4o}{\textbf{GPT-4o}} 
& N/D & \ding{51} & OpenAI ToU \\
\href{https://platform.openai.com/docs/models/gpt-5.5}{\textbf{GPT-5.5}} 
& N/D & \ding{51} & OpenAI ToU \\
\href{https://ai.google.dev/gemini-api/docs/models#gemini-3.5-flash}{\textbf{Gemini-3.5-Flash}} 
& N/D & \ding{51} & Google API ToS \\
\bottomrule
\end{tabular}
\caption{Overview of language models used in experiments. 
Prop.: proprietary status. N/D: not disclosed.}
\label{tab:models}
\end{table*}


\begin{table}[t!]
  \centering
  {
  \begin{tabular}{lr}
    \toprule
    \textbf{Country} & \multicolumn{1}{c}{\textbf{Cohen's $\kappa$ score}} \\
    \midrule
    EN  & 0.64 \\
    CN     & 0.55 \\
    KR      & 0.79 \\
    ES      & 0.63 \\
    ID     & 0.68 \\
    \bottomrule
  \end{tabular}
  }
  \caption{Inter-reviewer agreement by country.}
  \label{tab:kappa-country}
\end{table}

\begin{table}[t!]
\centering
\resizebox{\linewidth}{!}
{
\begin{tabular}{l|rrr}
\toprule
\multicolumn{1}{c}{\textbf{Model}} & \multicolumn{1}{c}{\textbf{High-sim}} & \multicolumn{1}{c}{\textbf{Low-sim}} & \multicolumn{1}{c}{\textbf{Gap (High - Low)}} \\
\midrule
Llama   & 49.66 & 65.96 & -16.30 \\
Mistral        & 51.23 & 57.03 & -5.80 \\
Qwen3       & 49.78 & 51.56 & -1.78 \\
GPT-3.5        & 59.82 & 67.41 & -7.59 \\
GPT-4o         & 76.01 & 80.35 & -4.35 \\
GPT-5o mini    & 69.39 & 78.83  & -9.44 \\
\bottomrule
\end{tabular}
}
\caption{
  Comparison of accuracy for high-similarity and low-similarity questions and performance gaps.
}
\label{tab:similarity-gap}
\end{table}

We evaluate six LLMs, including the closed-source models 
GPT-4o~\cite{openai2024gpt4technicalreport}, 
GPT-5.5~\cite{singh2026openaigpt5card}, and Gemini-3.5-flash~\cite{google2026gemini35flash} as well as the open-source models Llama-3.1-8B-Instruct~\cite{meta2024llama31}, 
Gemma-3-12B-IT~\cite{gemma_2025}, and 
Qwen3-8B~\cite{qwen3technicalreport} 
(see Table~\ref{tab:models}).
All models were evaluated using comparable prompt formats and 
decoding parameters (temperature=0.0, top\_p=0.95, 
max\_completion\_tokens=256) to ensure comparability and 
reproducibility.
GPT-5.5 does not support temperature control. 
We therefore use the default setting (temperature=1).
Gemini-3.5-Flash uses an internal thinking mode that 
consumes output tokens for reasoning. We increase 
max\_completion\_tokens to 2048 for this model to ensure 
complete responses.

Each model performs three independent inference runs 
corresponding to the three question types: Explicit Awareness 
Questions (binary Yes/No), Implicit Compliance Questions 
(5-choice MCQ), and Knowledge Verification Questions 
(5-choice MCQ).
Each run is single-pass with no fine-tuning or model 
modification.
GPT-4o and GPT-5.5 are accessed via the OpenAI API, and 
Gemini-3.5-Flash is accessed via the Google AI 
API. Open-source models are served locally with greedy decoding.

Experiments were executed on a server equipped with four 
NVIDIA RTX~A6000 GPUs (48GB each) and an Intel(R) Xeon(R) 
Gold~5218R CPU (2.10\,GHz, 256\,GB RAM).
Open-source models were each served on a single GPU using 
up to two GPUs simultaneously.
On the local environment, each open-source model completed 
all three evaluation stages within approximately 2 hours.
Closed-source models required 1--2 hours per model due to 
API overhead.
Overall computational time across all models amounted to 
approximately 12 GPU hours.

All predictions were stored in JSON files with raw model 
responses, parsed answers, and correctness labels.
Accuracies were computed by counting correct responses for 
each analysis dimension (stage, boundary type, region, and 
model).
All models and libraries were properly cited according to 
their respective licenses.



\begin{table}[t!]
\centering
\small
\begin{tabular}{lc}
\toprule
\textbf{Setting} & \textbf{BC Acc. (Cross-Univ.)} \\
\midrule
Independent & 10.4\% (76/732) \\
Sequential  & 80.6\% (590/732) \\
\midrule
Difference  & $+70.2$\%p \\
\bottomrule
\end{tabular}
\caption{
BC accuracy under independent versus sequential BA{\textrightarrow}BC measurement
(GPT-4o, Cross-Universe, $N=732$).
In the sequential setting, the model answers the BA question and then the
corresponding BC question within the same conversation.
}
\label{tab:sequential}
\end{table}

\begin{table}[t!]
\centering
\small
\begin{tabular}{lccc}
\toprule
\textbf{Model} & \textbf{KVQ\%} & \textbf{NR\% (No-role)} & \textbf{Both\%} \\
\midrule
Gemini-3.5-Flash & 100.0 & 99.5 & 99.5 \\
GPT-5.5          & 98.3  & 95.6 & 94.8 \\
GPT-4o           & 78.3  & 93.3 & 75.0 \\
Gemma-3-12B      & 80.8  & 84.2 & 68.4 \\
Llama-3.1-8B     & 87.2  & 77.0 & 68.7 \\
Qwen3-8B         & 50.5  & 64.4 & 29.7 \\
\bottomrule
\end{tabular}
\caption{
Factual accuracy on Cross-Universe questions under role change (KVQ) and role
removal (No-role). \textbf{KVQ\%}: correct as the target character;
\textbf{NR\%}: correct with no role assignment; \textbf{Both\%}: correct under
both conditions. All rates exceed the 20\% chance level of a five-choice MCQ.
}
\label{tab:norole}
\end{table}

\begin{table}[t!]
\centering
\small
\resizebox{\columnwidth}{!}{%
\begin{tabular}{l|rr|rr}
\toprule
\textbf{Region} & \textbf{C-Gap} & $n$ & \textbf{Override} & $n$ \\
\midrule
EN (US/UK) & $48.6 \pm 20.9$ & 244 & $90.8 \pm 12.2$ & 124 \\
Spain & $56.0 \pm 21.1$ & 280 & $78.2 \pm 30.2$ & 160 \\
China & $45.7 \pm 20.8$ & 280 & $83.3 \pm 19.5$ & 160 \\
Indonesia & $37.9 \pm 15.8$ & 248 & $83.2 \pm 23.3$ & 128 \\
Korea & $36.5 \pm 11.8$ & 280 & $68.9 \pm 30.6$ & 160 \\
\bottomrule
\end{tabular}
}%
\caption{Compliance Gap rate (across both boundary types) and parametric override rate (Cross-Universe only) by region, averaged across the six models (mean $\pm$ std). The two $n$ columns give the instances underlying each rate: all Compliance-Gap items (Temporal + Cross-Universe) and Cross-Universe items, respectively.}
\label{tab:cultural_uncertainty}
\end{table}

\begin{table}[t!]
\centering
\small
\resizebox{\columnwidth}{!}{%
\begin{tabular}{ll|rr}
\toprule
\textbf{Model} & \textbf{Region} & \textbf{C-Gap} & \textbf{Override} \\
\midrule
GPT-4o & EN & 79.5 & 100.0 \\
       & Spain & 90.4 & 70.6 \\
       & China & 75.7 & 83.3 \\
       & Indonesia & 65.7 & 57.1 \\
       & Korea & 49.6 & 80.0 \\
\midrule
GPT-5.5 & EN & 38.9 & 100.0 \\
        & Spain & 58.2 & 92.9 \\
        & China & 58.2 & 100.0 \\
        & Indonesia & 43.5 & 100.0 \\
        & Korea & 50.4 & 100.0 \\
\midrule
Gemini-3.5-Flash & EN & 40.6 & 100.0 \\
                 & Spain & 43.6 & 100.0 \\
                 & China & 31.1 & 100.0 \\
                 & Indonesia & 29.0 & 100.0 \\
                 & Korea & 24.3 & 100.0 \\
\midrule
Llama-3.1-8B & EN & 63.5 & 91.3 \\
             & Spain & 66.8 & 95.3 \\
             & China & 50.7 & 81.2 \\
             & Indonesia & 39.5 & 100.0 \\
             & Korea & 39.3 & 35.0 \\
\midrule
Gemma-3-12B & EN & 49.6 & 83.3 \\
            & Spain & 47.9 & 90.3 \\
            & China & 42.1 & 88.0 \\
            & Indonesia & 22.6 & 50.0 \\
            & Korea & 24.6 & 68.2 \\
\midrule
Qwen3-8B & EN & 19.7 & 70.0 \\
         & Spain & 29.3 & 20.0 \\
         & China & 16.4 & 47.1 \\
         & Indonesia & 27.0 & 92.3 \\
         & Korea & 30.7 & 30.4 \\
\bottomrule
\end{tabular}
}%
\caption{Full model$\times$region breakdown of Compliance Gap rate (\%, across both boundary types) and parametric override rate (\%, Cross-Universe only).}
\label{tab:region_model_detail}
\end{table}

%% file: 6-5-appendix_prompt.tex
\paragraph{Prompt for Boundary-Awareness (BA) Stage.}
Table~\ref{tab:ba_prompt} shows the prompt used for Explicit 
Awareness Questions. The model is asked to answer in binary 
Yes/No.

\paragraph{Prompt for Boundary-Compliance (BC) Stage.}
Table~\ref{tab:bc_prompt} shows the prompt used for Implicit 
Compliance Questions and Knowledge Verification Questions. 
Both use the same MCQ format. Only the role character and 
question content differ.

\paragraph{Prompt for Generating Multiple Choice Options.}
Table~\ref{tab:multiple_choice_prompt} shows the two-stage 
prompt used to generate distractors for compliance questions.

\begin{table*}[p!]
    \scriptsize
    \centering
\begin{tabular}{@{}p{\linewidth}@{}}
\toprule
\textbf{Boundary-Awareness (BA) Prompt} \\
\midrule
{
You are \{character\}. Please answer the following question 
as \{character\}.

~

And there is a \{character\} profile:

\{profile\}.

~

Question: \{question\}

~

Answer only ``Yes'' or ``No''.

Answer:
}\vspace{0.5ex}\\
\bottomrule
\end{tabular}
    \caption{Prompt for the Boundary-Awareness (BA) stage. 
    The model answers in binary Yes/No format.}
    \label{tab:ba_prompt}
\end{table*}

\begin{table*}[p!]
    \scriptsize
    \centering
\begin{tabular}{@{}p{\linewidth}@{}}
\toprule
\textbf{Boundary-Compliance (BC) / Knowledge Verification 
(KV) Prompt} \\
\midrule
{
You are \{character\}. Please read the following 
multiple-choice question and select the correct answer. 
Your response should be made as if you are \{character\}, 
and you must choose the answer that \{character\} would 
most likely provide.

~

And there is a \{character\} profile:

\{profile\}.

~

Now select the correct answer.

Question: \{question\}

Choices:

1. \{option1\}

2. \{option2\}

3. \{option3\}

4. \{option4\}

5. \{option5\}

~

First, briefly explain your reasoning. Then, state your 
final answer on the last line in the format: Answer: 
<number>
}\vspace{0.5ex}\\
\bottomrule
\end{tabular}
    \caption{Prompt for the Boundary-Compliance (BC) and 
    Knowledge Verification (KV) stages. For BC, \{character\} 
    is the role character and the correct answer is the 
    abstention option. For KV, \{character\} is the target 
    character and the correct answer is the factually 
    correct option.}
    \label{tab:bc_prompt}
\end{table*}

\begin{table*}[p!]
    \scriptsize
    \centering
\begin{tabular}{@{}p{\linewidth}@{}}
\toprule
\textbf{First Stage for generating multiple choice options} \\
\midrule
{
You are an uncertainty benchmark generator.

~

Your task is to generate four incorrect answers that are 
similar to the correct answer.

~

Requirements:

You must be 100\% certain that each generated answer is 
factually incorrect.

Start from the correct answer and create four alternative 
answers by making minimal edits (up to 15 words).

Each answer must be plausible but clearly wrong.

~

The question is directed to \{profile\}, and the answer 
should be written as if \{profile\} himself is responding.

~

Please return only the incorrect answers in the following 
format:

Incorrect Answer 1: [text]

Incorrect Answer 2: [text]

Incorrect Answer 3: [text]

Incorrect Answer 4: [text]

~

Now generate four incorrect answers for the following 
question.

Question: \{Question\}

Correct answer: \{Answer\}
}\vspace{0.5ex}\\
\midrule
\textbf{Second Stage for generating multiple choice options} \\
\midrule
{
You are an uncertainty benchmark generator.

Your task is to generate four additional incorrect answers 
that do not overlap with the given four incorrect answers.

~

Requirements:

You must be 100\% certain each new answer is factually 
incorrect.

All answers must be plausible but incorrect.

Each of the four new incorrect answers (Incorrect Answer 5 
to 8) must use entirely different sentence structures from 
Incorrect Answers 1 to 4, avoiding similarity in syntax, 
phrase order, or grammatical patterns to ensure maximum 
variety and distinctiveness.

~

The question is directed to \{profile\}, and the answer 
should be written as if \{profile\} himself is responding.

~

Please return only the new incorrect answers in the 
following format:

Incorrect Answer 5: [text]

Incorrect Answer 6: [text]

Incorrect Answer 7: [text]

Incorrect Answer 8: [text]

~

Now generate new incorrect answers that do not overlap 
with the following:

Question: \{Question\}

Correct answer: \{Answer\}

Incorrect answer1: \{Incorrect1\}

Incorrect answer2: \{Incorrect2\}

Incorrect answer3: \{Incorrect3\}

Incorrect answer4: \{Incorrect4\}
}\vspace{0.5ex}\\
\bottomrule
\end{tabular}
    \caption{Two-stage prompt for generating multiple choice 
    distractors.}
    \label{tab:multiple_choice_prompt}
\end{table*}

\begin{table*}[p!]
    \scriptsize
    \centering
\begin{tabular}{@{}p{\linewidth}@{}}
\toprule
\textbf{Criteria 1:} Whether the correct answer is indeed the most appropriate and accurate response to the question\\
\quad \textbullet\; Score 1 if the correct answer is appropriate; otherwise, score 0\\[2pt]

\textbf{Criteria 2:} Whether the incorrect answers are not mistakenly interpreted as correct\\
\quad \textbullet\; Score 1 if the incorrect answers are clearly not correct; score 0 if any are essentially the same as the correct answer\\[2pt]

\textbf{Criteria 3:} Whether there are any overlaps in meaning among the incorrect answers\\
\quad \textbullet\; Score 1 if there is no semantic overlap; score 0 if there is redundancy or repetition\\[2pt]

\textbf{Criteria 4:} Whether the answers are relevant to the question\\
\quad \textbullet\; Score 1 if relevant; score 0 if any answer is unrelated to the question\\
\bottomrule
\end{tabular}
\caption{Validation Criteria}
\label{tab:val_criteria}
\end{table*}

%% file: 7-6-appendix_human.tex
We recruit two native reviewers from each region, all born, 
raised, and educated in their respective countries, to validate 
all questions and answers.
Reviewers assess question appropriateness, verify answer 
correctness, and evaluate distractor distinctness and 
plausibility.
Items flagged by at least one reviewer undergo verification 
and revision, with inter-reviewer disagreements resolved 
through consensus. Unresolved items are removed.

The validation process consists of three steps:
\begin{enumerate}
    \item \textbf{Criteria-based validation.} All native 
    reviewers examine question-answer pairs using the four 
    predefined validation criteria in 
    Table~\ref{tab:val_criteria}.
    \item \textbf{Linguistic and cultural appropriateness.} 
    Reviewers verify that each item is both linguistically 
    and culturally correct. Items flagged by any reviewer 
    are re-evaluated by other reviewers.
    \item \textbf{Semantic distinctiveness.} Similar 
    distractors within a question are filtered out and 
    regenerated to prevent quality degradation.
\end{enumerate}


We also consider cultural variation across countries, as 
the multi-stage validation sometimes results in conflicts 
between reviewers. In such cases, both reviewers and the 
authors re-examine the item to reach a consensus. Questions 
that do not reach consensus are removed to maintain benchmark 
reliability.

To quantify inter-reviewer reliability, we measure Cohen's 
$\kappa$ for each country. As shown in 
Table~\ref{tab:kappa-country}, the kappa scores indicate a 
generally substantial level of agreement across all regions.

%% file: 7-8-appendix_similarity.tex

To assess the plausibility of the CHARM distractors, we conducted an additional analysis examining the semantic similarity between each distractor and the correct answer, as shown in Table \ref{tab:similarity-gap}. We computed cosine similarity between the correct answer and each distractor using the paraphrase-multilingual-MiniLM-L12-v2 \cite{reimers-2019-sentence-bert} embedding model, which provides a consistent, model-agnostic metric across closed/open-LLMs. We then compared model performance on the top 20\% (high-similarity) and bottom 20\% (low-similarity) questions. These results demonstrate that semantically close distractors consistently increase the benchmark’s difficulty.

%% file: 7-9-appendix_sequentialExp.tex
\label{app:sequential}

In the main experiments, BA and BC are measured in independent runs.
To assess how shared context affects the recognition--compliance dissociation, we additionally run a sequential condition in which the model first answers the BA question and then receives the corresponding BC question within the same conversation (GPT-4o, Cross-Universe, $N=732$).


BC accuracy increases from 10.4\% to 80.6\% (Table~\ref{tab:sequential}). This substantial increase suggests that the model's preceding ``No'' response strongly influences subsequent refusal behavior. Consequently, the sequential setting does not allow us to disentangle genuine boundary compliance from conversational consistency with the model's prior response, and we therefore do not attribute the improvement to boundary compliance alone.

Two conclusions follow.
First, the independent design avoids this consistency confound and yields a conservative lower bound on compliance capacity. The large gap that remains under this stricter measurement supports interpreting the dissociation as a genuine recognition--compliance gap.
Second, the large improvement in the sequential setting suggests a simple mitigation direction: explicitly eliciting the character's knowledge boundary before answering can substantially increase subsequent abstention. However, further work is needed to determine the extent to which this improvement reflects genuine boundary compliance rather than consistency bias in multi-turn, interactive settings.

%% file: 7-10-appendix_noRole.tex
\label{app:norole}

The Knowledge Verification Question (KVQ) establishes parametric access by re-posing the question to the target character.
As a complementary control that removes the role entirely, we additionally pose the same Cross-Universe BC questions with no role assignment (No-role).
If the model answers correctly without any role, the relevant fact is accessible independently of the role-change manipulation.

Table~\ref{tab:norole} reports both verifications, and two observations follow.
First, the two verifications converge: No-role accuracy (64.4--99.5\%) confirms that the models possess the relevant knowledge without any role, consistent with the KVQ rates (50.5--100.0\%), so evidence from role change (KVQ) and role removal (No-role) agrees.
Second, a conservative estimate requiring both verifications (Both\%) still ranges 29.7--99.5\%, well above the 20\% chance level of a five-choice MCQ.
Together, these controls strengthen the interpretation that compliance failures reflect accessible parametric knowledge rather than coincidental correctness.

%% file: 7-11-appendix_per-regionUncertaintyandItemCounts.tex
\label{app:region_detail}
Table~\ref{tab:cultural_uncertainty} reports, for each region, the Compliance Gap and parametric override rates averaged across the six models with standard deviation, together with the number of underlying instances.
Table~\ref{tab:region_model_detail} provides the full model$\times$region breakdown.
While regional differences are visible across models, their magnitude varies (e.g., override: Korea $68.9 \pm 30.6$, Spain $78.2 \pm 30.2$), so we read these patterns as broad tendencies rather than fixed cultural properties.